\documentclass{article}

\usepackage{PRIMEarxiv}
\usepackage[utf8]{inputenc} 
\usepackage[T1]{fontenc}    
\usepackage{hyperref}       
\usepackage{url}            
\usepackage{amsmath}
\usepackage{algorithm}
\usepackage{algorithmic}
\usepackage{booktabs}       
\usepackage{amsfonts}       
\usepackage{nicefrac}       
\usepackage{microtype}      
\usepackage{lipsum}
\usepackage{fancyhdr}       
\usepackage{graphicx}  
\graphicspath{{media/}}     
\usepackage{filecontents}
\usepackage{multirow} 
\usepackage{caption}

\usepackage{amssymb}
\usepackage{amsthm}
\usepackage{mathtools}
\usepackage{tikz}
\usetikzlibrary{arrows.meta,positioning}
\usepackage{listings}
\usepackage{xcolor}

\title{Behavioral Governance for Autonomous AI Agents: The AgentBound Framework}

\author{
Anuj Kaul$^{1}$, Qianlong Lan$^{1}$, Pranay Gupta$^{2}$ \\
$^{1}$ eBay Inc. \\
$^{2}$ Independent Researcher
}

\begin{document}
\maketitle

\begin{abstract}
Autonomous AI agents increasingly perform consequential actions on behalf of human principals, including financial transactions, external communications, and enterprise workflows. Existing agent infrastructure relies on identity federation and delegated authorization to authenticate workloads and control resource access, but it cannot determine whether an authorized action should be executed under the current behavioral and operational context.

We present AgentBound, a runtime governance framework that provides verifiable behavioral oversight for autonomous AI agents. AgentBound evaluates each proposed action using three independent authorities: delegated authorization, owner-signed behavioral constitutions, and site action contracts. Their judgments are conservatively composed through a formal decision model to determine whether an action should be permitted, reviewed, or denied before execution.

To provide accountability, AgentBound generates cryptographically verifiable governance receipts that bind every action to the exact delegation, policy, and semantic artifacts governing the decision, enabling independent replay verification and policy provenance. The framework also introduces standing delegation for long-running agents, allowing periodic workloads to operate under continuously refreshed governance policies while preserving revocability and bounded authority.

We present the formal foundation, system architecture, governance receipt protocol, and AgentBound-Bench, a benchmark framework for evaluating governance correctness, authority composition, and accountability. Rather than replacing model alignment, AgentBound complements it by providing a deterministic governance layer between authorization and execution, transforming governance from a process that must be trusted into one that can be independently verified.

\end{abstract}


\section{Introduction}

Autonomous AI agents are rapidly evolving from conversational assistants into delegated actors capable of executing consequential actions on behalf of human principals. Modern agent workloads routinely browse websites, access enterprise infrastructure, manage communications, schedule events, invoke external APIs, and coordinate complex, multi-step workflows without continuous human oversight~\cite{voix2025,savaglio2022governance}. As agent autonomy increases and execution becomes increasingly decoupled from real-time human intervention, establishing comprehensive behavioral oversight emerges as a first-class systems challenge rather than a mere application-layer concern.

Existing security and governance infrastructures primarily address two foundational questions: identity systems establish \emph{who} the agent is, while delegated authorization frameworks determine \emph{what} resources or endpoints the agent is permissioned to access~\cite{spiffe,rfc8693}. However, these traditional access-control mechanisms remain fundamentally blind to a third, critical axis of operational security: \emph{Should a particular action occur under the current, dynamic runtime circumstances?} 

An agent may be flawlessly authenticated, operating within its valid OAuth or IAM scope, and fully compliant with static access policies, yet still execute decisions that directly violate its principal's intent. Such behavioral deviations include executing premature financial transactions, publishing unreviewed public content, leaking sensitive information under adversarial context manipulation, or selecting irreversible actions when safe alternatives exist. These systemic failures stem neither from identity flaws nor authorization deficiencies; they represent a fundamental failure of behavioral governance.

\begin{figure*}[ht]
\centering
\includegraphics[width=\textwidth]{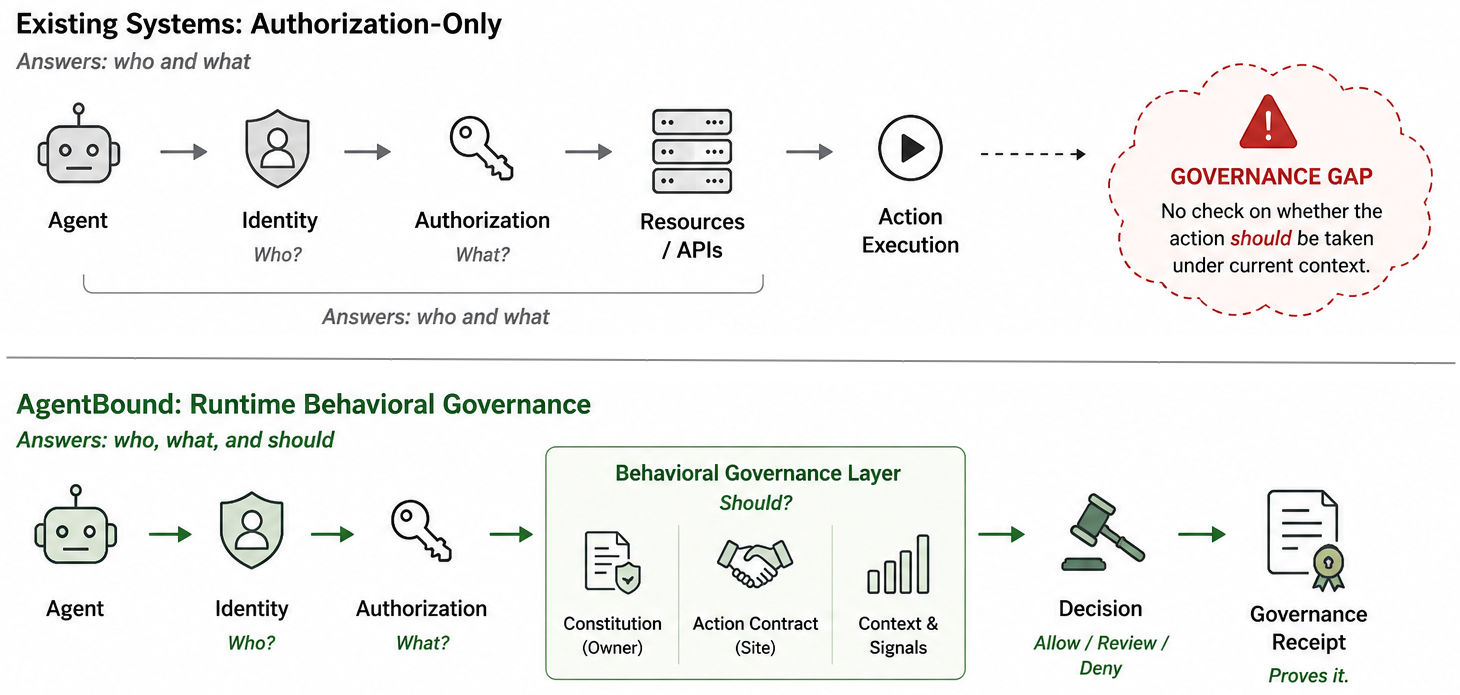}
\caption{The paradigm shift from access authorization to verifiable behavioral governance. Existing systems resolve agent identity and resource entitlements but leave a critical execution blind spot regarding context suitability. AgentBound introduces a non-bypassable behavioral governance layer between authorization and execution, synthesizing multi-authority policies into a deterministic verdict accompanied by a cryptographically signed governance receipt.}
\label{fig:governance-gap}
\end{figure*}

This governance gap becomes acute as agents transition to persistent, long-running systems. Current architectures lack formal mechanisms to express nuanced owner intent, enforce behavioral boundaries at execution time, or produce non-repudiable evidence of policy compliance~\cite{wang2026agentspec,bhardwaj2026abc,zhang2026history,auditableagents2026}. Crucially, this behavioral governance deficit cannot be resolved by training-time model alignment alone. While alignment techniques attempt to influence a model's internal probabilistic reasoning, runtime governance enforces deterministic, context-aware constraints over proposed actions against evolving operational, organizational, and environmental realities~\cite{bai2022constitutional}.

To bridge this gap, we present AgentBound, a runtime governance framework that provides verifiable behavioral oversight for autonomous AI agents. As illustrated in Figure~\ref{fig:governance-gap}, AgentBound intercepts the agent's execution trajectory by introducing an explicit governance layer immediately following authorization. AgentBound can be summarized by three questions:

\begin{itemize}
\item Who is the agent?
\item What may the agent access?
\item Should this action occur?
\end{itemize}

The framework evaluates proposed actions across three independent pillars of authority: delegated authorization, owner-signed behavioral constitutions, and site-specific action contracts. The judgments emitted by these distinct authorities are formally composed via a conservative decision algebra to yield a definitive, non-bypassable enforcement verdict before any action touches an underlying system.

Beyond strict runtime enforcement, AgentBound introduces the architecture of \emph{governance receipts}. These receipts are cryptographically verifiable artifacts that tightly bind each executed action to the exact snapshot of behavioral policies, structural delegations, and semantic site contracts responsible for the decision~\cite{aegon2026}. Unlike conventional, passive audit logs, governance receipts feature replay-verifiable provenance. This allows independent third-party auditors to deterministically reconstruct and validate policy compliance post-hoc, without relying on the internal state or the runtime environment of the agent. Consequently, AgentBound shifts agent accountability from an unverifiable model of implicit trust into a framework of transparent, independent verification.

In summary, this paper makes the following four primary contributions:
\begin{itemize}
\item \textbf{Three-Authority Governance Model:} We formalize a comprehensive model that unifies delegated authorization, owner behavioral constitutions, and site action contracts to evaluate the contextual validity of proposed agent actions.
\item \textbf{Standing Delegation Architecture:} We design a secure, long-running delegation mechanism tailored for periodic autonomous agents that preserves policy freshness and granular revocability.
\item \textbf{Formal Runtime Enforcement:} We introduce a deterministic enforcement pipeline wherein complementary judgments from independent authorities are synthesized through a conservative decision algebra prior to execution.
\item \textbf{Cryptographic Governance Receipts:} We develop a verifiable accountability mechanism that produces replay-verifiable provenance, enabling independent post-hoc auditability of autonomous agent operations.
\end{itemize}

\section{Motivation and Governance Gap}

\subsection{Identity and Authorization Are Not Governance}

The emergence of autonomous AI agents has accelerated the development of identity frameworks (such as SPIFFE/SPIRE) \cite{spiffe} designed to authenticate software entities operating on behalf of human users. While these systems effectively resolve workload identity to establish downstream accountability, identity alone remains fundamentally decoupled from the behavioral correctness of the actions performed by that agent. Similarly, modern authorization architectures predicated on OAuth scopes, JWT claims, capability tokens, and structured policy engines are designed to restrict the boundary of resources an agent may access. While authorization effectively constrains the operational perimeter of a workload, it remains blind to whether a particular action should be performed under a given set of runtime circumstances. For example, an agent may possess valid delegated OAuth scopes to issue financial refunds up to a specific monetary threshold. A request to execute a refund within that authorized limit falls entirely within the access control baseline; however, standard authorization engines cannot determine whether that specific decision demands manager intervention, requires explicit customer consent, meets internal confidence thresholds, or violates seasonal operational blackout periods. Authorization fails to address these critical boundary conditions because they pertain to behavioral intent and contextual compliance rather than to resource accessibility.

\subsection{Behavioral Governance Failures and the Structural Gap}

We formalize a behavioral governance failure as a scenario in which an autonomous agent operates entirely within its permitted authorization boundaries yet violates the strategic or behavioral expectations imposed by its human principal. These systemic failures manifest through various patterns, such as publishing externally visible corporate content without mandatory editorial reviews, executing high-stakes financial transactions below acceptable model confidence thresholds, or opting for irreversible system mutations when equivalent reversible alternatives are available. 

The execution lifecycle of an autonomous agent action traditionally progresses through three distinct operational phases: authentication, authorization, and execution. A critical architectural gap persists between the authorization and execution phases. Current systems evaluate whether an agent has the technical capability to perform an action, but they rarely evaluate whether that action aligns with owner-defined behavioral expectations under current ambient conditions. Human principals delegate transactional authority to autonomous entities for operational efficiency, yet they retain implicit behavioral and risk-mitigation expectations that cannot be expressed through rigid access-control lists or role-based primitives. We define this missing structural layer as behavioral governance. AgentBound is designed specifically to bridge this architectural gap by introducing deterministic runtime behavioral oversight and cryptographically verifiable accountability mechanisms into the agent execution pipeline.

\section{Threat Model}

\subsection{Principals and Adversarial Objective}

We formalize the AgentBound ecosystem around four core architectural principals, denoted as the tuple $\mathcal{P} = (O, A, AB, S)$:
\begin{itemize}
    \item \textbf{Owner ($O$):} The human principal who deploys an agent, authorizes structural delegation, and defines the underlying behavioral governance requirements through a ground-truth intent profile $\mathcal{I}_O$.
    \item \textbf{Agent ($A$):} The autonomous software system commissioned to act on behalf of the owner. $A$ generates a sequence of proposed actions $\alpha \in \mathcal{A}$ based on its internal execution state. While trusted to execute assigned tasks within its technical boundaries, $A$ is not trusted to correctly govern its own behavior or evaluate policy compliance.
    \item \textbf{AgentBound Service ($AB$):} The centralized runtime governance enforcement engine serving as a trusted computing base (TCB). $AB$ intercepts each proposed action $\alpha$, evaluates it against the joint multi-authority policy $\mathcal{M}$, and outputs a deterministic enforcement verdict $v \in \{\textsf{Allow}, \textsf{Deny}\}$.
    \item \textbf{Site Resource Provider ($S$):} The external application, web platform, or remote enterprise service that hosts the target resources and executes the validated actions.
\end{itemize}

Let $\mathcal{R}_S(\alpha) \in \{\textsf{Valid}, \textsf{Invalid}\}$ represent the standard access control and delegation verification function of the site $S$ under a delegated authorization scope $\mathcal{G}_{\text{auth}}$. We formalize the adversarial environment by considering a behavioral adversary $\mathcal{A}_{\text{beh}}$ capable of interacting with the agent's runtime environment, input streams, or prompt space. 

Unlike conventional privilege-escalation adversaries who aim to bypass $\mathcal{R}_S$ such that an unauthorized action is executed, the objective of the behavioral adversary $\mathcal{A}_{\text{beh}}$ is to manipulate or induce an authorized agent $A$ into executing actions that directly violate the Owner's intent while remaining strictly within the agent's delegated permission scope. Formally, given a proposed action $\alpha$ generated under the influence of $\mathcal{A}_{\text{beh}}$, the attack is successful if and only if the following joint condition holds:
\begin{equation}
\mathcal{R}_S(\alpha) = \textsf{Valid} \quad \land \quad \alpha \notin \mathcal{I}_O.
\end{equation}
The adversary exploits cognitive ambiguity, semantic uncertainty, or incomplete governance rule specifications to achieve this state, inducing an agent to misallocate funds without obtaining required administrative approvals, tricking it into manipulating internal model confidence metrics, or forcing unauthorized external publication of sensitive data, all while maintaining valid OAuth tokens or cryptographic session credentials.

\subsection{System Boundaries and Trust Assumptions}

AgentBound focuses specifically on mitigating semantic and behavioral threats that manifest after traditional authorization has successfully concluded. The framework seeks to systematically detect and intercept runtime behavioral policy violations, unauthorized execution bypassing mandatory approval workflows, unsafe confidence-dependent decisions induced by anomalous environments, and direct violations of site action semantics or reversibility constraints. 

The security guarantees of AgentBound are predicated on a clear set of foundational trust assumptions. We assume that the cryptographic signing keys belonging to both the Owner and the AgentBound enforcement service remain uncompromised throughout the execution lifecycle. We further assume that the runtime environment successfully guarantees that all governance evaluations occur strictly prior to the dispatch of any consequential action to an external system. Finally, we assume that all generated governance receipts are securely transmitted to and stored within an append-only, tamper-evident audit ledger.

To clarify the operational boundaries of this framework, certain categories of agent security vulnerabilities are explicitly considered out of scope. AgentBound is not designed to defend against adversarial prompt injection attacks, direct large language model jailbreaks, baseline model alignment failures, or hardware-level side-channel data exfiltration. Consequently, AgentBound must be conceptualized as an administrative governance and behavioral accountability layer rather than a totalizing agent security architecture.

\section{AgentBound Formal Foundation and Governance Model}

The architectural layout of AgentBound is dictated by three core design tenets: the separation of governance authorities, behavioral accountability, and replay verifiability. Rather than modeling agent cognition, planning, or internal reasoning loops, this section formally defines how governance decisions are represented, composed, and verified.

\subsection{Canonical Actions and Governance Judgments}

To enforce uniform policy evaluation across heterogeneous environments, AgentBound abstracts all runtime events into a structured representation termed a canonical action. Formally, a canonical action is a tuple defined as:
\[
a = (\text{operation}, \text{resource}, \text{parameters}, \text{risk}, \text{context})
\]
where $\text{operation}$ identifies the semantic intent of the invocation; $\text{resource}$ denotes the target object; $\text{parameters}$ encompass action-specific arguments; $\text{risk}$ encapsulates the declared exposure category; and $\text{context}$ contains volatile execution metadata. Each independent governance authority processes this canonical action to yield a structured governance judgment, formalized as the tuple:
\[
J = (\text{verdict}, \text{constraints}, \text{obligations}, \text{provenance})
\]
where $\text{verdict}$ designates the evaluation outcome; $\text{constraints}$ define bounded execution thresholds; $\text{obligations}$ denote mandatory runtime dependencies (such as human approvals or individual-level data redacting); and $\text{provenance}$ cryptographically maps the judgment to its originating policy version.

\subsection{The Three Pillars of Authority}

The evaluation framework partitions oversight across three decoupled semantic dimensions. Delegated authorization operates as an access-control authority, determining structural entitlement by verifying resource perimeters, spending boundaries, and API token scopes to answer whether the agent is broadly permitted to execute a category of action. The behavioral constitution functions as a principal-centric authority, mapping owner-authored rules, confidence metrics, and escalation criteria to evaluate whether the proposed action aligns with user expectations under present ambient conditions. The site action contract serves as an interface-centric authority, validating remote risk classification, visibility vectors, and structural reversibility to declare what the execution fundamentally alters on the target system. 

\begin{figure*}[ht]
\centering
\includegraphics[width=\textwidth]{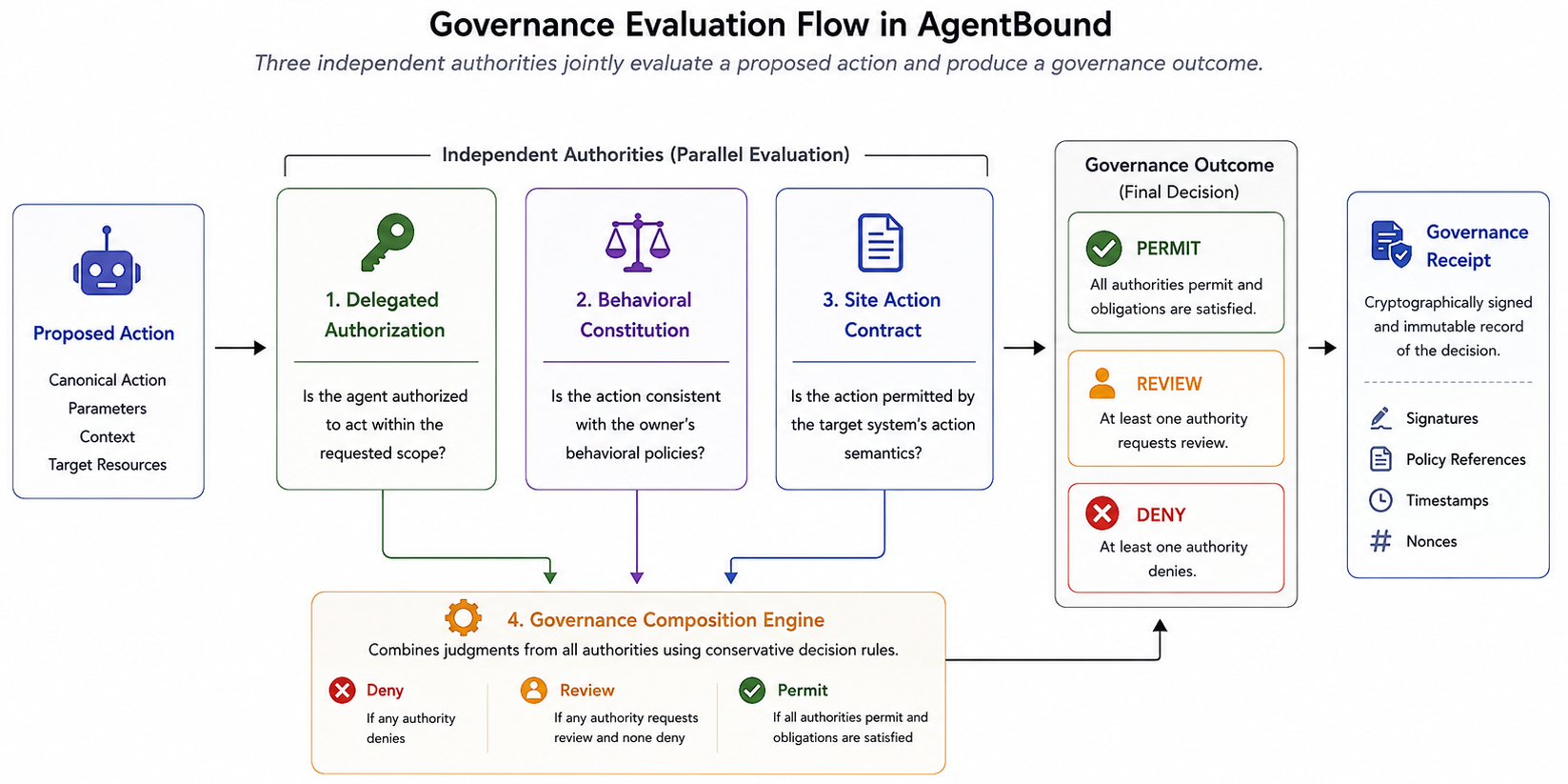}
\caption{Governance evaluation flow in AgentBound. A proposed action is independently evaluated by three authorities: delegated authorization, behavioral constitution, and site action contract. Their judgments are conservatively composed to produce a final governance outcome, associated obligations, and a signed governance receipt.}
\label{fig:governance-flow}
\end{figure*}

As illustrated in Figure~\ref{fig:governance-flow}, the runtime engine routes a canonical action through parallel evaluation pipelines corresponding to each authority. The intermediate judgments are then collected by the composition engine, which resolves conflicts using a conservative decision algebra.

\subsection{Decision Composition and Execution Semantics}

The decision space maps to three discrete outcomes ordered by a lattice of strict logical restrictiveness:
\[
\text{Deny} < \text{Review} < \text{Permit}
\]
The composition engine unifies the parallel judgment vectors into a singular runtime directive governed by four monotonic invariants: the most restrictive verdict dominates the lattice, constraints aggregate conjunctively, obligations accumulate cumulatively, and structural provenance records must be preserved intact. Under these rules, an isolated denial cannot be overridden, review thresholds cannot be bypassed, and no authority can implicitly expand the perimeter established by another. 

An action is granted physical execution clearance if and only if the final composed decision evaluates to permit and all accumulated runtime obligations are fully satisfied. Satisfying a review obligation satisfies a behavioral policy requirement but does not alter or expand the underlying authorization boundary. Consequently, an action blocked by delegated authorization cannot be validated or made executable by subsequent behavioral approvals. This algebraic composition yields four fundamental system properties: authorization preservation, behavioral independence, obligation separation, and replay verifiability.

\section{System Architecture}

\begin{figure}[t]
\centering
\includegraphics[width=\linewidth]{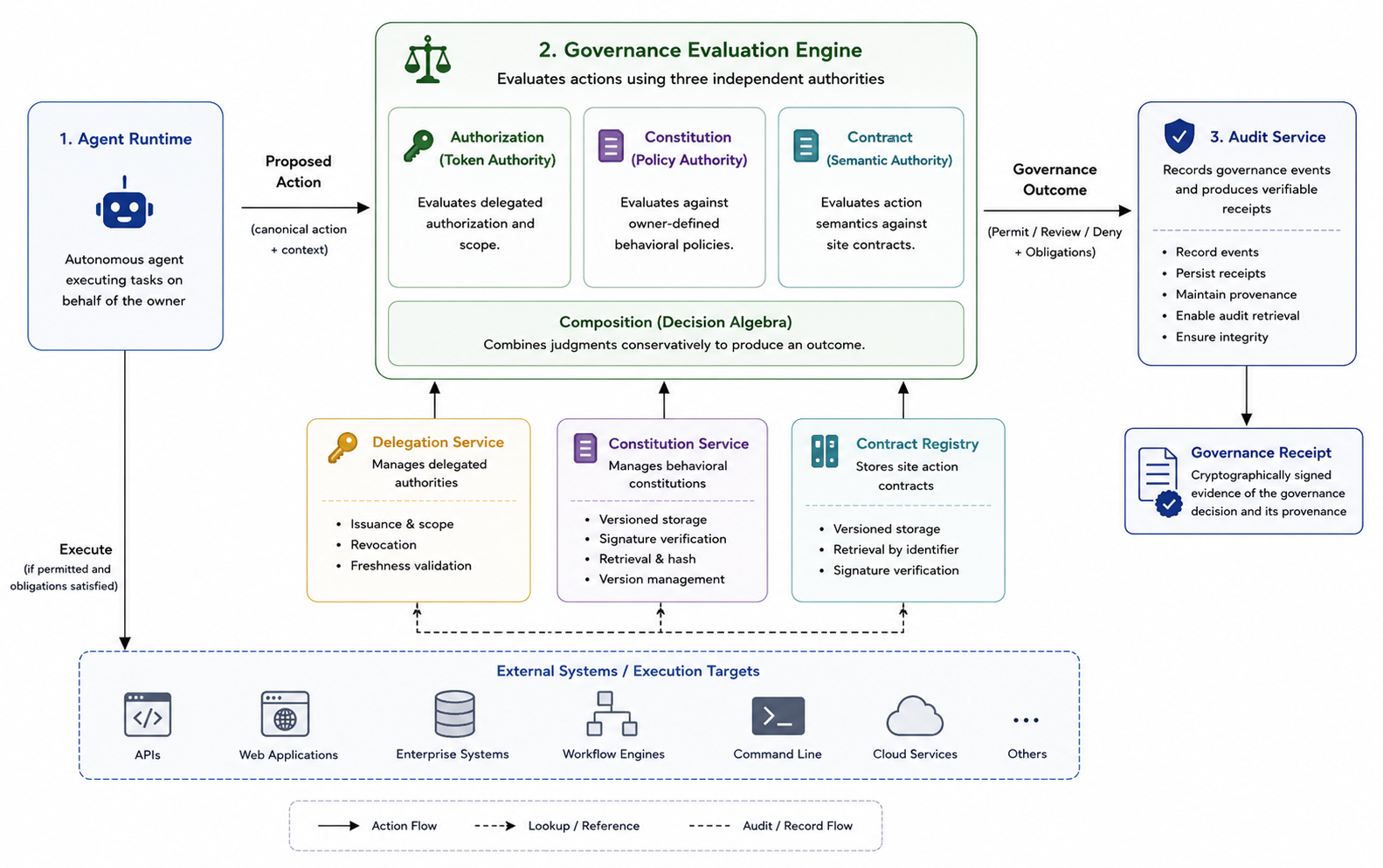}
\caption{High-level AgentBound architecture.
Agent-generated actions are evaluated by a governance engine that combines delegated authorization, behavioral constitutions, and site action contracts before execution. Governance outcomes and provenance records are persisted through cryptographically verifiable governance receipts and an audit service.}
\label{fig:architecture}
\end{figure}

AgentBound is implemented as a decoupled, runtime governance layer detached from agent decision-making and positioned directly before action execution. The architecture is guided by the core invariant that every consequential action must pass through an external governance checkpoint prior to target system mutation. Figure~\ref{fig:architecture} illustrates the structural relationship between these system components.

The framework unifies five distinct functional components into a non-bypassable governance pipeline that sequentially evaluates, logs, and validates agent behaviors.

\subsection{Core Components and Services}

The Agent Runtime encapsulates the execution environment performing tasks on behalf of a human principal. From a security posture perspective, the runtime is explicitly treated as untrusted; AgentBound assumes that client-side agent reasoning may deviate from owner intent due to jailbreaks, thereby requiring out-of-band validation prior to system execution. The Delegation Service manages the structural scope of authorized capability issued to the agent, handling token issuance, structural revocation, and delegation freshness validation. 

The Constitution Service acts as the authoritative repository for owner-defined behavioral constraints. Each constitution is maintained as a versioned, cryptographically signed, and immutable policy artifact that is indexed and retrievable via a unique cryptographic hash reference. The Governance Evaluation Engine functions as the core policy enforcement arbiter of the architecture. Upon receiving an action request, it validates delegation token freshness, resolves the active constitution version, fetches the matching site action contract, unifies their intermediate judgments via our formal decision algebra, and commits the execution verdict. Crucially, the engine evaluates policy isolated from side effects, ensuring the complete absence of state mutations on external endpoints during the evaluation cycle. Finally, the Audit Service provides the observability substrate required for system-wide accountability and non-repudiation by appending governance events and persisting signed receipts.

\section{Governance Receipts Architecture}

AgentBound addresses the limitations of traditional audit logs by introducing governance receipts, defined as cryptographically verifiable, immutable primitives that bind each discrete agent action to the exact policy artifacts responsible for its runtime outcome. Rather than functioning as passive event streams, these receipts encapsulate the joint state of evaluated authorities, enabling independent, out-of-band replay verification long after execution has concluded.

\begin{figure*}[t]
\centering
\includegraphics[width=\textwidth]{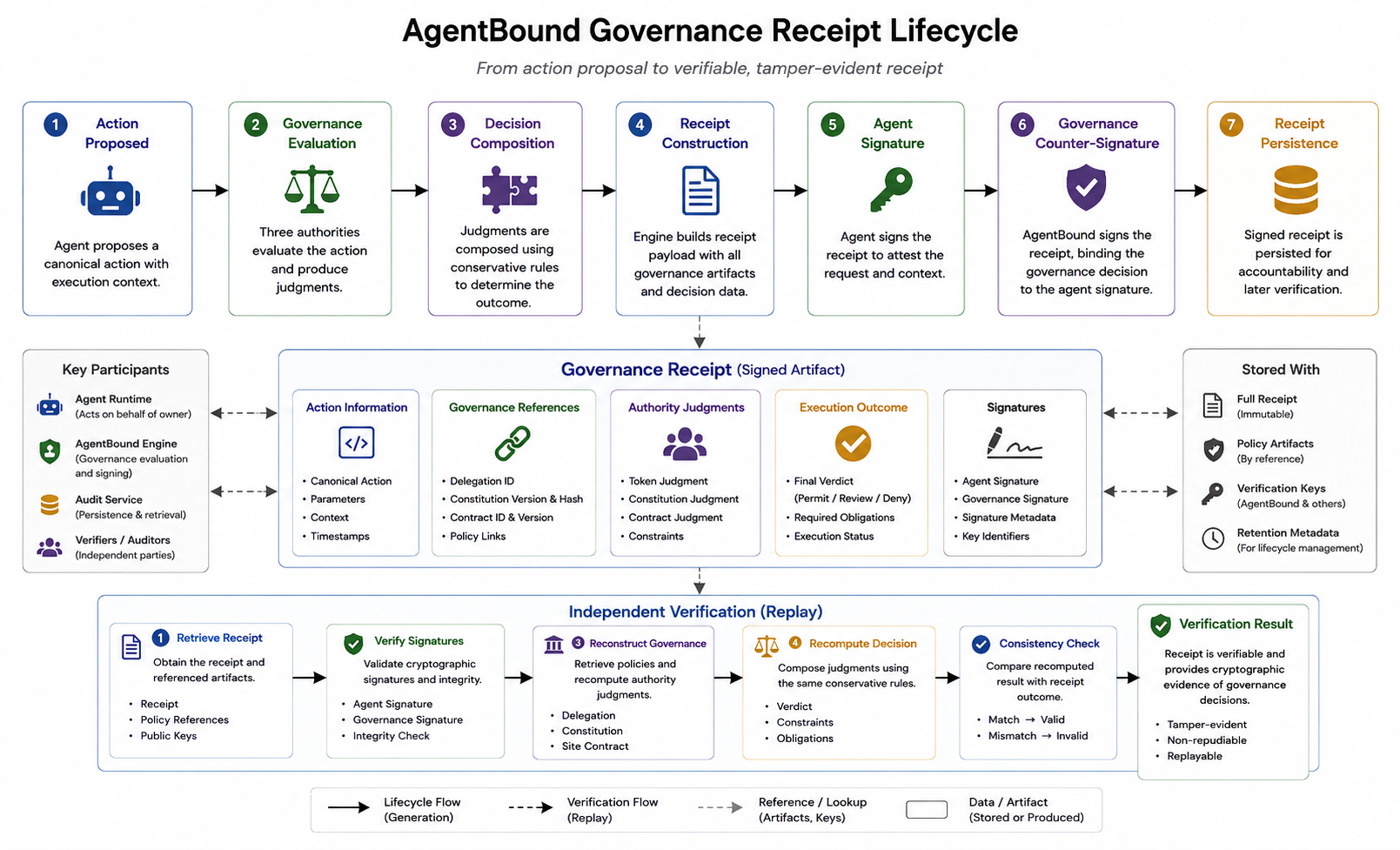}
\caption{Governance receipt lifecycle in AgentBound. A governance receipt is generated after action evaluation, signed by both the agent and the AgentBound governance service, persisted for accountability, and later independently verified through replay using referenced governance artifacts and public verification keys.}
\label{fig:receipt-lifecycle}
\end{figure*}

As illustrated in Figure~\ref{fig:receipt-lifecycle}, the lifecycle model guarantees that receipts are synthesized synchronously upon the conclusion of policy evaluation, co-signed through a non-interactive ordered protocol by both the requesting agent and the enforcement service, persisted to an append-only audit ledger, and made available for public replay validation.

\subsection{Receipt Structure and Pipeline}

Formally, a governance receipt is modeled as the tuple:
\[
\text{Receipt} = (\text{Action}, \text{Governance}, \text{Judgments}, \text{Outcome}, \text{Signatures})
\]
The $\text{Action}$ component encapsulates the invariant fields of the canonical action. The $\text{Governance}$ block establishes non-repudiable policy provenance by capturing the explicit delegation token identifier, the behavioral constitution version number along with its cryptographic hash, and the active site action contract version identifier. The $\text{Judgments}$ category materializes the individual, typed evaluations, while the $\text{Outcome}$ segment registers the final operational primitive. The $\text{Signatures}$ block embeds the asymmetric cryptographic metadata required to guarantee global integrity and non-repudiation.

Receipt generation proceeds via a deterministic four-stage pipeline. Following evaluation, the system constructs the raw receipt payload. Accountability binding is subsequently achieved through a two-phase ordered countersigning protocol. First, the requesting agent signs the constructed receipt payload using its private asymmetric key. Second, the AgentBound enforcement service intercepts this token and countersigns the entire block, covering the core receipt payload, the agent's signature, and the terminal decision vector. Because the service's signature encloses the agent's cryptographic signature, any out-of-band modification to either the underlying action payload or the intermediate governance evaluation invalidates the verification chain.

The verification architecture enables external, untrusted entities to validate policy compliance asynchronously. Verifying a receipt requires only the artifact payload, the corresponding public verification keys, and the historical governance artifacts fetched via the receipt's stored references. The validation sequence comprises cryptographic integrity verification, governance reconstruction, and a deterministic decision consistency check that independently replays the canonical action against the reconstructed policy models. This layout yields five core security invariants: integrity, non-repudiation, policy provenance, replay verifiability, and decision transparency.

\begin{table}[htbp]
\centering
\caption{Example Governance Receipt}
\label{tab:example-receipt}
\begin{tabular}{ll}
\toprule
Field & Value \\
\midrule
Receipt ID & r-1024 \\
Agent ID & ads-agent \\
Action & launch campaign \\
Delegation Ref & d-118 \\
Constitution Version & 3 \\
Token Judgment & Permit \\
Constitution Judgment & Review \\
Final Decision & Review \\
Required Obligation & Human Approval \\
\bottomrule
\end{tabular}
\end{table}

\section{Standing Delegation for Persistent Agents}

\begin{figure*}[t]
\centering
\includegraphics[width=\textwidth]{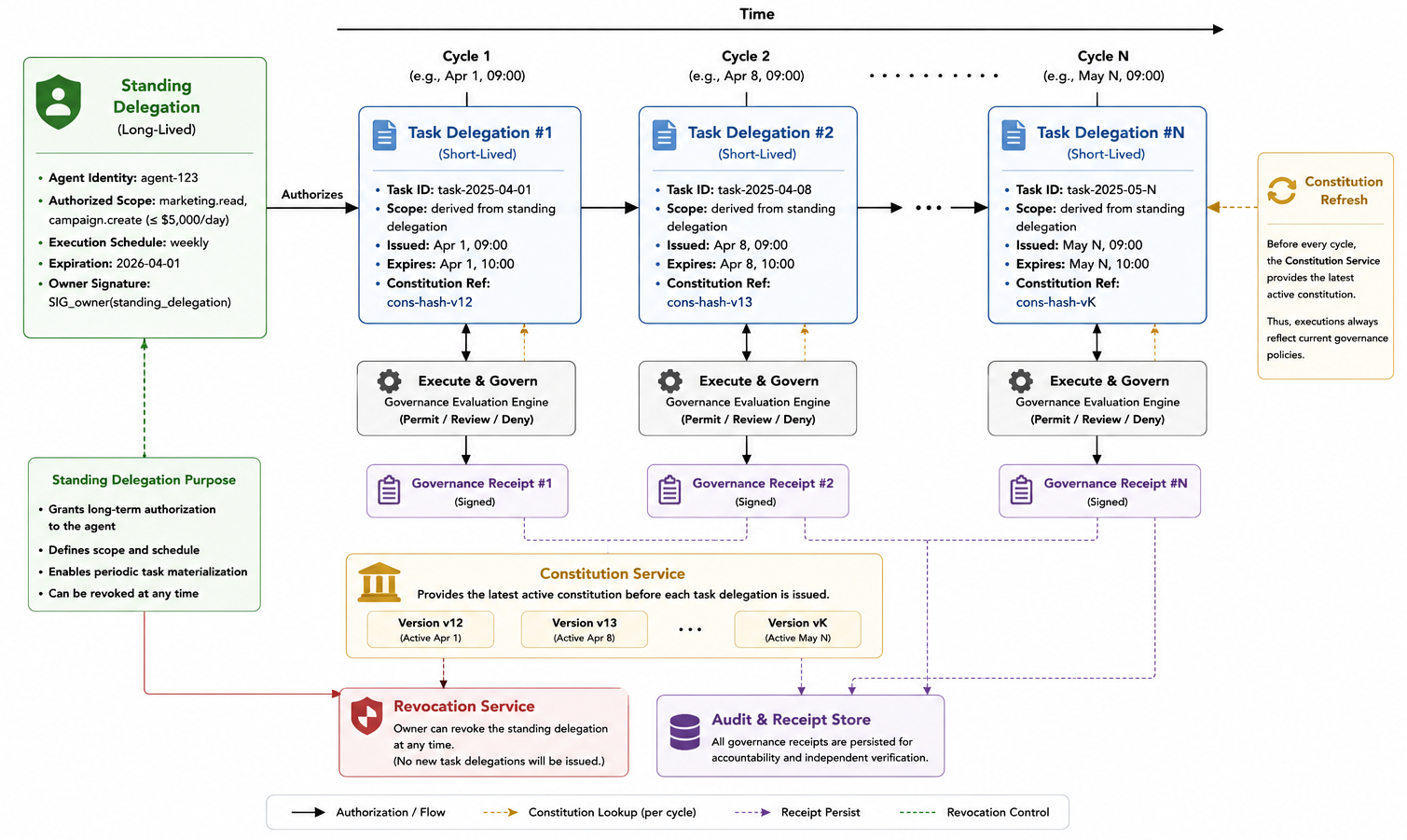}
\caption{The AgentBound standing delegation model. A long-lived standing delegation authorizes the creation of short-lived task delegations for each execution cycle. Before every cycle, the Constitution Service provides the latest active constitution, ensuring that periodic agents automatically operate under current governance policies while maintaining revocability and limited exposure.}
\label{fig:standing-delegation}
\end{figure*}

Many autonomous agents operate persistently without continuous human supervision, executing recurrent or long-running tasks asynchronously over extended operational horizons. Traditional delegation and access control frameworks operate under the implicit assumption that a human principal is actively present or within a valid session context when a transaction is initiated. This assumption fails to hold for periodic or long-running agentic workloads. AgentBound addresses this structural limitation by introducing a standing delegation model that cleanly decouples long-term user authorization from short-lived runtime execution authority. As illustrated in Figure~\ref{fig:standing-delegation}, the framework isolates these concern vectors through standing delegations paired with per-cycle task materialization.

\subsection{The Periodic Agent Problem and Two-Level Model}

Standard authorization infrastructure is predominantly predicated on session-oriented tokens, which introduces severe systemic vulnerabilities for periodic autonomous workloads: no human user is present to evaluate ambient risks when an execution cycle spontaneously begins, and organizational governance policies may have shifted significantly during the interval between discrete execution cycles. 

To mitigate these risks, AgentBound formalizes a two-level cryptographic delegation architecture. The foundational tier is the standing delegation, which represents the long-term strategic authorization granted explicitly by the human owner. It encapsulates the cryptographically bound agent identity, the maximum permissible operational scope, the scheduled execution interval, a hard expiration epoch, and the owner's verification signature. Standing delegations are intentionally restricted from direct execution capabilities within target environments; instead, they serve exclusively as a root of authority empowering the runtime engine to dynamically materialize short-lived task delegations.

\subsection{Task Materialization and Automatic Constitution Refresh}

At the initialization boundary of every discrete execution cycle, the runtime engine evaluates the active standing delegation to materialize an ephemeral task delegation. This task-level credential is derived strictly from the parent standing delegation boundaries and encapsulates a unique task identifier, a contextually constrained operational scope, an explicit issuance timestamp, a short-lived expiration threshold, and an immutable reference to the governing constitution. Task delegations are designed to be explicitly short-lived, minimizing the window of vulnerability associated with token leakage.

A core characteristic of this architecture is the deterministic synchronization of active policy state via an automatic constitution refresh mechanism. Prior to the materialization of a task delegation, the runtime requests the current active constitution version from the Constitution Service. The newly generated task token encapsulates the precise hash reference of this latest policy artifact. Consequently, modifications enacted by the owner automatically propagate to subsequent execution cycles without necessitating manual agent reauthorization or human intervention. This model introduces four fundamental governance benefits and operational invariants: revocability, policy freshness, limited exposure, and governance continuity.

\section{Evaluation Framework}
This section defines the evaluation methodology and benchmark design. Empirical results will be reported in a future release.
Evaluating governance systems differs fundamentally from evaluating autonomous agent architectures. While conventional benchmarks typically quantify task completion rates, core reasoning quality, planning token efficiency, or tool-use success, they are designed primarily to assess whether an agent can successfully achieve an operational goal. AgentBound addresses a fundamentally orthogonal axis, focusing instead on whether a runtime governance mechanism can correctly evaluate, constrain, and explain consequential actions. To support the systematic benchmarking of such defensive layers, we introduce AgentBound-Bench, a standardized evaluation framework designed specifically to quantify governance enforcement correctness, multi-authority composition behavior, and policy provenance integrity.

\begin{figure*}[ht]
\centering
\includegraphics[width=\textwidth]{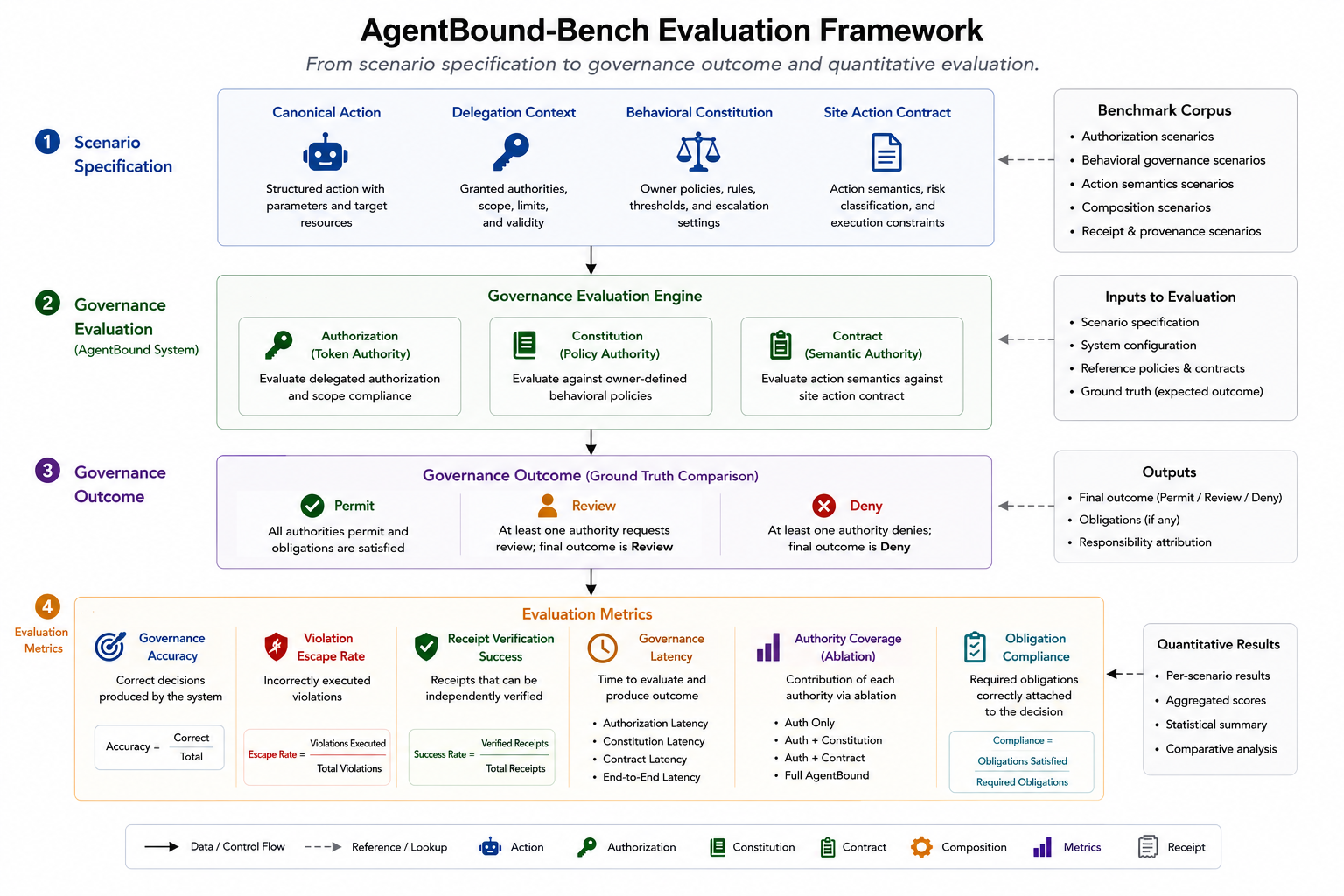}
\caption{AgentBound-Bench evaluation framework. Each benchmark scenario combines a canonical action, delegation context, behavioral constitution, and site action contract. Governance outcomes are compared against ground truth to evaluate enforcement correctness, authority composition behavior, receipt verification, and operational performance.}
\label{fig:agentbound-bench}
\end{figure*}

As illustrated in Figure~\ref{fig:agentbound-bench}, AgentBound-Bench isolates the assessment of governance infrastructure from the underlying language model's cognitive capabilities. Rather than evaluating task optimization, this testing framework explicitly measures whether the runtime architecture safely constrains and cryptographically validates malicious or anomalous agent actions against known ground-truth behaviors.

\subsection{Evaluation Objectives and Benchmark Structure}

The empirical assessment of AgentBound-Bench is structured around four interconnected operational objectives: governance enforcement correctness, authority composition behavior, accountability and provenance, and operational performance. Together, these vectors capture both the security effectiveness and the practical enterprise deployability of the framework.

Every benchmark scenario within AgentBound-Bench is formalized as a four-component tuple consisting of a canonical action payload, a specific structural delegation context, a signed behavioral constitution version, and a remote site action contract. Given this invariant input matrix, the expected governance verdict is deterministically pre-defined, encapsulating the expected intermediate authority judgments, the terminal outcome directive, all mandatory runtime obligations, and the specific authority responsible for the root decision. This un-entangled design decouples policy evaluation from the non-deterministic state space of underlying language models, enabling perfectly reproducible and uniform security testing.

\subsection{Scenario Taxonomy, Metrics, and Ablation}

The test suite partition encompasses five specialized scenario categories tailored to stress-test discrete segments of the governance layer: authorization scenarios, behavioral governance scenarios, action semantics scenarios, composition scenarios, and receipt scenarios. 

AgentBound-Bench quantifies framework efficacy through four core mathematical and system metrics. Governance accuracy measures the statistical fraction of evaluation scenarios where the runtime engine produces a verdict identical to the formal ground truth, expressed as:
\[
\text{Accuracy} = \frac{\text{Correct Decisions}}{\text{Total Decisions}}
\]
To evaluate the absolute security boundary, the violation escape rate quantifies the precise fraction of policy-violating agent actions that bypass enforcement blocks and reach physical execution targets, formalized as:
\[
\text{Escape Rate} = \frac{\text{Violations Executed}}{\text{Total Violations}}
\]
where minimizing this value indicates stronger runtime enforcement boundaries. Cryptographic accountability is captured via the receipt verification success rate, while system overhead is monitored through fine-grained governance latency.

To empirically isolate the distinct security coverage contributed by each independent governance authority, each scenario undergoes multi-stage ablation execution across four baseline configurations: authorization validation in isolation, authorization paired with constitutional policies, authorization synthesized with site action contracts, and the full multi-authority AgentBound deployment pipeline. This framework establishes the formal benchmark specifications necessary for rigorous validation; future work will document comprehensive empirical results across large-scale synthetic datasets.

\section{Related Work}

AgentBound positions itself at the critical intersection of autonomous agent governance, runtime compliance enforcement, cryptographic delegation provenance, and verifiable system auditability. 

\subsection{Agent Governance and Functional Verification}

The problem of constraining autonomous agent behaviors at runtime has catalyzed several recent enforcement paradigms. AgentSpec establishes a customizable runtime enforcement layer for large language model agents through a dedicated, rule-based domain-specific language that accommodates both preventive and corrective operational modalities~\cite{wang2026agentspec}. Addressing probabilistic bounds, Agent Behavioral Contracts leverages formal programmatic contracts characterized by strict drift limits and statistical satisfaction guarantees~\cite{bhardwaj2026abc}. Similarly, Agent Contracts investigates resource-bounded foundational contracts targeted at cooperative multi-agent architectures~\cite{bosello2026contracts}. AgentBound diverges from these frameworks by treating behavioral constitutions as owner-defined, cryptographically signed, and highly versioned immutable policy artifacts that are composed via a formal decision algebra and verified through signed governance receipts.

Our formulation of a behavioral constitution also relates to, but departs from, optimization paradigms in model alignment and constitutional learning. Constitutional AI pioneered the use of natural language principles to guide training-time algorithmic alignment~\cite{bai2022constitutional}. Extending this paradigm to multi-agent environments, Multi-Agent Constitutional Learning automates the discovery of constitutional architectures directly from historical execution trajectories~\cite{thareja2026mac}. AgentBound departs from these approaches by enforcing compliance deterministically at the system runtime boundary rather than optimizing model weights or internal reasoning tendencies at training time, optimizing for verifiable system accountability rather than statistical behavioral shaping.

\subsection{Audit, Provenance, and Authorization Infrastructure}

Verifiable auditability forms another critical pillar of related literature. The Right to History architecture introduces tamper-evident agent execution records designed to establish a sovereign kernel for verifiable agent histories~\cite{zhang2026history}. From a mediation cost perspective, Auditable Agents develops structured evidence-design tenets to systematically optimize performance trade-offs associated with pre-execution runtime inspection~\cite{auditableagents2026}. To reinforce data access controls, Aegon implements hardware-attested compliance mechanisms that safeguard content extraction via ledger-bound token architectures and cryptographically signed mobile receipts~\cite{aegon2026}. AgentBound shares these high-level accountability objectives but embeds precise hashes of the delegation scope, constitution version, and site contract state active at the moment of evaluation, allowing independent third parties to completely reconstruct the original governance decision pipeline.

In the context of large-scale distributed architectures, industrial reference implementations have heavily shaped formal authorization primitives. Google Zanzibar introduces a globally consistent, fine-grained relationship-based authorization engine~\cite{zanzibar2019}. Similarly, Open Policy Agent decouples security logic through a general-purpose declarative policy language~\cite{opa}, while Cedar provides a highly specialized authorization grammar optimized for fine-grained delegation boundaries within distributed enterprise networks~\cite{cedar2023}. These production frameworks focus primarily on entitlement validation, answering whether a specific authenticated entity possesses technical permissions to access a targeted resource. AgentBound establishes authorization as one governance input among a parallel matrix of independent arbiters to evaluate contextual invariants that cannot be expressed via traditional resource-level access control primitives.

Finally, our standing delegation architecture builds upon established cryptographic token exchange and human-to-agent delegation lineage models. The Human Delegation Provenance protocol designs a lightweight cryptographic routine that explicitly anchors downstream terminal tool executions to a root human principal across complex, multi-hop agentic delegation chains~\cite{dalugoda2026hdp}. Structurally, RFC 8693 establishes the industry standard for OAuth 2.0 Token Exchange, providing formal semantic foundations for programmatic token transformation and delegation chain security within distributed environments~\cite{rfc8693}. AgentBound extends these concepts by injecting a runtime layer of behavioral constraint directly on top of token validation and dynamically materializing short-lived, ephemeral task tokens from an unexpired, long-term authorization root.

\section{Discussion and Limitations}

AgentBound is designed as a modular runtime governance and behavioral accountability framework for autonomous agents. As agent capabilities expand across interconnected digital infrastructure, behavioral governance increasingly emerges as a distinct system concern that cannot be conceptualized as a sub-problem of identity federation, access control, or training-time model alignment. 

\subsection{System Boundaries and Structural Comparisons}

AgentBound is structurally decoupled from traditional, perimeter-centric security engineering frameworks. Conventional system security infrastructure focuses primarily on mitigating unauthorized resource access, privilege escalation, data compromise, and malicious code injection. Conversely, AgentBound addresses situations where a fully authenticated and legitimately authorized agent executes actions that directly violate its principal's intent while remaining strictly within its delegated access control scope. Because these anomalies constitute behavioral governance failures rather than access control breaches, AgentBound must be conceptualized as a complementary system layer that operates alongside traditional network controls.

While access control mechanisms dictate whether an agent possesses the structural entitlement to access a remote resource, governance determines whether a specific canonical action should occur under a given set of runtime environmental variables. A robust authorization engine can successfully restrict the perimeter of accessible endpoint APIs, yet it remains fundamentally incapable of expressing contextual behavioral variables, such as dynamic model confidence thresholds, mandatory administrative approval chains, or temporal organizational blackout windows. Consequently, AgentBound does not replace authorization systems; instead, it establishes delegated authorization as a single, foundational input to be processed within a broader, multi-authority decision algebra.

Furthermore, AgentBound diverges fundamentally from optimization-centric model alignment methods. Alignment shapes model cognition beforehand; governance evaluates the final behavior before system mutation. This relationship is directly analogous to the software engineering boundary separating application correctness from localized operational policy. Runtime governance cannot substitute for training-time alignment, as an enforcement engine does not alter how a model plans or interprets instructions, focusing exclusively on intercepting actions prior to remote dispatch.

Finally, existing logging frameworks provide runtime telemetry and tracing capabilities for agent execution, yet they typically record only what occurred during system mutation without providing a verifiable derivation explaining why a specific decision was permitted, blocked, or escalated. AgentBound extends traditional observability by generating atomic governance receipts. These receipts encapsulate exact policy version states, delegation token identifiers, individual authority judgments, terminal outcomes, and asymmetric cryptographic signatures, allowing auditors to perform historical replay verification and completely reconstruct the original decision algebra out-of-band.

\subsection{Current Limitations and Future Directions}

The security guarantees of AgentBound are bounded by several foundational limitations. First, the framework assumes a strict operational sequence where governance evaluation occurs non-bypassably prior to execution; any agent workload capable of directly muting external endpoints without passing through the enforcement layer falls outside the system boundary. Second, AgentBound isolates behavior at the action perimeter rather than filtering cognitive model inputs, rendering it incapable of directly intercepting prompt injection, jailbreaks, or internal adversarial model manipulation. Third, the efficacy of the enforcement layer remains completely dependent on the precision and quality of the underlying constitutions and site action contracts. Fourth, the model relies on a trusted computing base, assuming that the core enforcement engine, signing keys, and policy registries remain completely uncompromised. Finally, comprehensive empirical validation across large-scale, automated enterprise agent workloads remains future work.

Several key architectural domains remain open for future exploration. Automated constitution generation presents an opportunity to assist administrators in distilling unstructured enterprise compliance guidelines into formalized, versioned behavior policies while preserving human-in-the-loop oversight. Research into adaptive governance could yield dynamic constitutions capable of updating enforcement constraints in response to observed agent drift without invalidating cryptographic provenance. Multi-party governance chapters are required to support federated environments involving overlapping stakeholders, such as enterprise deployers, infrastructure providers, and sovereign regulators. Finally, establishing interoperable governance receipt standards represents a critical path toward unified accountability across heterogeneous multi-agent ecosystems.

\section*{Artifact Availability}

To foster reproducible research and support open-source development within the autonomous agent governance domain, the reference implementation of AgentBound, alongside its constitution schema specifications and formalized governance receipt layouts, will be made publicly available under an open-source license. The associated evaluation framework, AgentBound-Bench, comprising compiled benchmark scenarios, baseline ground-truth annotations, out-of-band receipt verification tooling, and deployment documentation, will similarly be released to the community upon the conclusion of our initial empirical evaluation campaign.

\section{Conclusion}

Autonomous AI agents are fundamentally transitioning from passive conversational assistants into high-stakes delegated workloads capable of executing consequential tasks on behalf of human principals. As these systems achieve greater operational autonomy, traditional identity verification and resource-level access control are no longer sufficient to secure their execution perimeters. Identity infrastructure establishes who an agent is, and authorization frameworks define what resources an agent may access; however, neither paradigm addresses the critical runtime challenge of determining whether a specific action should occur under given ambient circumstances.

To address this structural gap, we presented AgentBound, a runtime governance framework that introduces verifiable behavioral oversight for autonomous systems. AgentBound integrates three independent governance authorities---delegated authorization, owner-signed behavioral constitutions, and site action contracts---into a unified, multi-authority decision engine. Under this paradigm, candidate tool invocations are evaluated non-bypassably before execution, intermediate judgments are composed conservatively via a formal decision algebra, and every discrete transaction is encapsulated within a cryptographically signed governance receipt. Unlike traditional, passive audit logging, these immutable receipts enable independent out-of-band decision reconstruction, shifting agent accountability from trusted platforms to absolute verification. 

Furthermore, the architecture introduces a two-level standing delegation model tailored to the lifecycle of persistent, periodic agents. This mechanism empowers long-running workloads to operate asynchronously under dynamically refreshed governance policies while preserving immediate user revocability and bounding credential exposure. AgentBound operates strictly at the system boundary as a deterministic enforcement layer positioned between authorization and execution; its objective is not to influence or interpret agent cognition, but to systematically govern how physical actions are evaluated, justified, and recorded. 

As autonomous workloads become deeply embedded within enterprise infrastructure, core digital services, and automated operational pipelines, behavioral governance will inevitably emerge as a foundational architectural requirement rather than an optional administrative feature. Future multi-agent ecosystems must demand rigorous behavioral oversight alongside baseline identity federation and token authorization. The operational safety of this paradigm is enforced by AgentBound's central architectural invariant: scope permitted it, constitution stopped it, and the receipt proves it. Ultimately, we hope this framework provides a practical and scalable blueprint for a future where autonomous AI workloads are not merely highly capable, but also structurally governable, accountable, and cryptographically verifiable.

\bibliographystyle{unsrt}  
\bibliography{references}  

@article{bai2022constitutional,
  title={Constitutional ai: Harmlessness from ai feedback},
  author={Bai, Yuntao and Kadavath, Saurav and Kundu, Sandipan and Askell, Amanda and Kernion, Jackson and Jones, Andy and Chen, Anna and Goldie, Anna and Mirhoseini, Azalia and McKinnon, Cameron and others},
  journal={arXiv preprint arXiv:2212.08073},
  year={2022}
}

@article{bhardwaj2026abc,
  title={Agent behavioral contracts: Formal specification and runtime enforcement for reliable autonomous AI agents},
  author={Bhardwaj, Varun Pratap},
  journal={arXiv preprint arXiv:2602.22302},
  year={2026}
}

@misc{bosello2026contracts,
      title={Agent Contracts: A Formal Framework for Resource-Bounded Autonomous AI Systems}, 
      author={Qing Ye and Jing Tan},
      year={2026},
      eprint={2601.08815},
      archivePrefix={arXiv},
      primaryClass={cs.MA},
      url={https://arxiv.org/abs/2601.08815}, 
}

@article{dalugoda2026hdp,
  title={HDP: A Lightweight Cryptographic Protocol for Human Delegation Provenance in Agentic AI Systems},
  author={Dalugoda, Asiri},
  journal={arXiv preprint arXiv:2604.04522},
  year={2026}
}

@article{savaglio2022governance,
  title={Governance of autonomous agents on the web: Challenges and opportunities},
  author={Kampik, Timotheus and Mansour, Adnane and Boissier, Olivier and Kirrane, Sabrina and Padget, Julian and Payne, Terry R and Singh, Munindar P and Tamma, Valentina and Zimmermann, Antoine},
  journal={ACM Transactions on Internet Technology},
  volume={22},
  number={4},
  pages={1--31},
  year={2022},
  publisher={ACM New York, NY}
}

@article{thareja2026mac,
  title={MAC: Multi-agent constitution learning},
  author={Thareja, Rushil and Gupta, Gautam and Pinto, Francesco and Lukas, Nils},
  journal={arXiv preprint arXiv:2603.15968},
  year={2026}
}

@inproceedings{wang2026agentspec,
  title={AgentSpec: Customizable runtime enforcement for safe and reliable LLM agents.(2026)},
  author={Wang, Haoyu and Poskitt, Christopher M and Sun, Jun},
  booktitle={Proceedings of the IEEE/ACM International Conference on Software Engineering, ICSE},
  pages={12--18},
  year={2026}
}

@article{zhang2026history,
  title={Right to History: A Sovereignty Kernel for Verifiable AI Agent Execution},
  author={Zhang, Jing},
  journal={arXiv preprint arXiv:2602.20214},
  year={2026}
}

@article{voix2025,
  title={Building the Web for Agents: A Declarative Framework for Agent-Web Interaction},
  author={Schultze, Sven and Kietzmann, Meike Verena and Sch{\"o}nfeld, Nils-Lucas and Stock-Homburg, Ruth},
  journal={arXiv preprint arXiv:2511.11287},
  year={2025}
}

@article{aegon2026,
  title={Aegon: Auditable AI Content Access with Ledger-Bound Tokens and Hardware-Attested Mobile Receipts},
  author={Baskaran, Amrish and Pherwani, Nirbhay and Krishnan, Raghul},
  journal={arXiv preprint arXiv:2604.06693},
  year={2026}
}

@article{auditableagents2026,
  title={Auditable agents},
  author={Nian, Yi and Yuan, Aojie and Zhang, Haiyue and Li, Jiate and Zhao, Yue},
  journal={arXiv preprint arXiv:2604.05485},
  year={2026}
}

@techreport{rfc8693,
  title={OAuth 2.0 token exchange},
  author={Jones, Michael and Nadalin, Anthony and Campbell, Brian and Bradley, John and Mortimore, Chuck},
  year={2020}
}

@inproceedings{zanzibar2019,
  title={Zanzibar:$\{$Google’s$\}$ Consistent, Global Authorization System},
  author={Pang, Ruoming and Caceres, Ramon and Burrows, Mike and Chen, Zhifeng and Dave, Pratik and Germer, Nathan and Golynski, Alexander and Graney, Kevin and Kang, Nina and Kissner, Lea and others},
  booktitle={2019 USENIX Annual Technical Conference (USENIX ATC 19)},
  pages={33--46},
  year={2019}
}

@article{opa,
  title={Ai-powered policy management: Implementing open policy agent (opa) with intelligent agents in kubernetes},
  author={Vadisetty, Rahul and Polamarasetti, Anand and others},
  journal={Cuestiones de Fisioterapia},
  volume={54},
  number={5},
  pages={19--27},
  year={2025}
}

@article{cedar2023,
  title={Cedar: A new language for expressive, fast, safe, and analyzable authorization},
  author={Cutler, Joseph W and Disselkoen, Craig and Eline, Aaron and He, Shaobo and Headley, Kyle and Hicks, Michael and Hietala, Kesha and Ioannidis, Eleftherios and Kastner, John and Mamat, Anwar and others},
  journal={Proceedings of the ACM on Programming Languages},
  volume={8},
  number={OOPSLA1},
  pages={670--697},
  year={2024},
  publisher={ACM New York, NY, USA}
}

@inproceedings{spiffe,
  title={SPIFFE-based zero-trust authentication for AI agent ecosystems},
  author={Pappu, Karthik and Bhushan, Badal and Mittal, Akshay},
  booktitle={2025 International Conference on Computer and Applications (ICCA)},
  pages={1--7},
  year={2025},
  organization={IEEE}
}

\end{document}